\pdfoutput=1
\documentclass{vgtc}
\usepackage{textcomp}
\usepackage{microtype}

\graphicspath{{figures/}{pictures/}{images/}{./}}

\usepackage{times}

\usepackage{tabu}
\usepackage{booktabs}
\usepackage{mwe}
\usepackage{mathptmx}
\usepackage{xspace}
\usepackage{tikz}

\onlineid{0}
\vgtccategory{Visualization and Rendering}
\vgtcinsertpkg

\renewcommand*{\backref}[1]{}

\title{Finetuning Prompt-Based Image Editor for Geometry-Consistent Room Defurnishing}

\author{Prakhar Kulshreshtha\thanks{e-mail: \{prakhar, bpugh, zachary.nussdorfer, salma\}@geomagical.com}
\and Brian Pugh\footnotemark[1]
\and Zachary Nussdorfer\footnotemark[1]
\and Salma Jiddi\footnotemark[1]}
\affiliation{\scriptsize Geomagical Labs, Inc.}

\teaser{
  \newlength{\teasercolwidth}
\setlength{\teasercolwidth}{0.22\linewidth}
\newcommand{\teaserimg}[2][]{\includegraphics[width=\teasercolwidth, alt={#1}]{#2}}

\centering
\footnotesize
\setlength{\tabcolsep}{3pt}

\newcommand{\teaserlegendline}[1]{\tikz[baseline=-0.5ex]{\draw[#1,opacity=0.5,line width=1.2pt] (0,0) -- (1.4em,0);}}
\begin{tikzpicture}
  \node[inner sep=0pt] (teasergrid) {%
    \begin{tabular}{@{}c@{\hspace{4pt}}c@{\hspace{4pt}}c@{\hspace{4pt}}c@{}}
      Input & SD + LoRA + Controlnet & FLUX.2 & FLUX.2 + LoRA (ours) \\[0.4em]
      \teaserimg[Room scan 1, input]{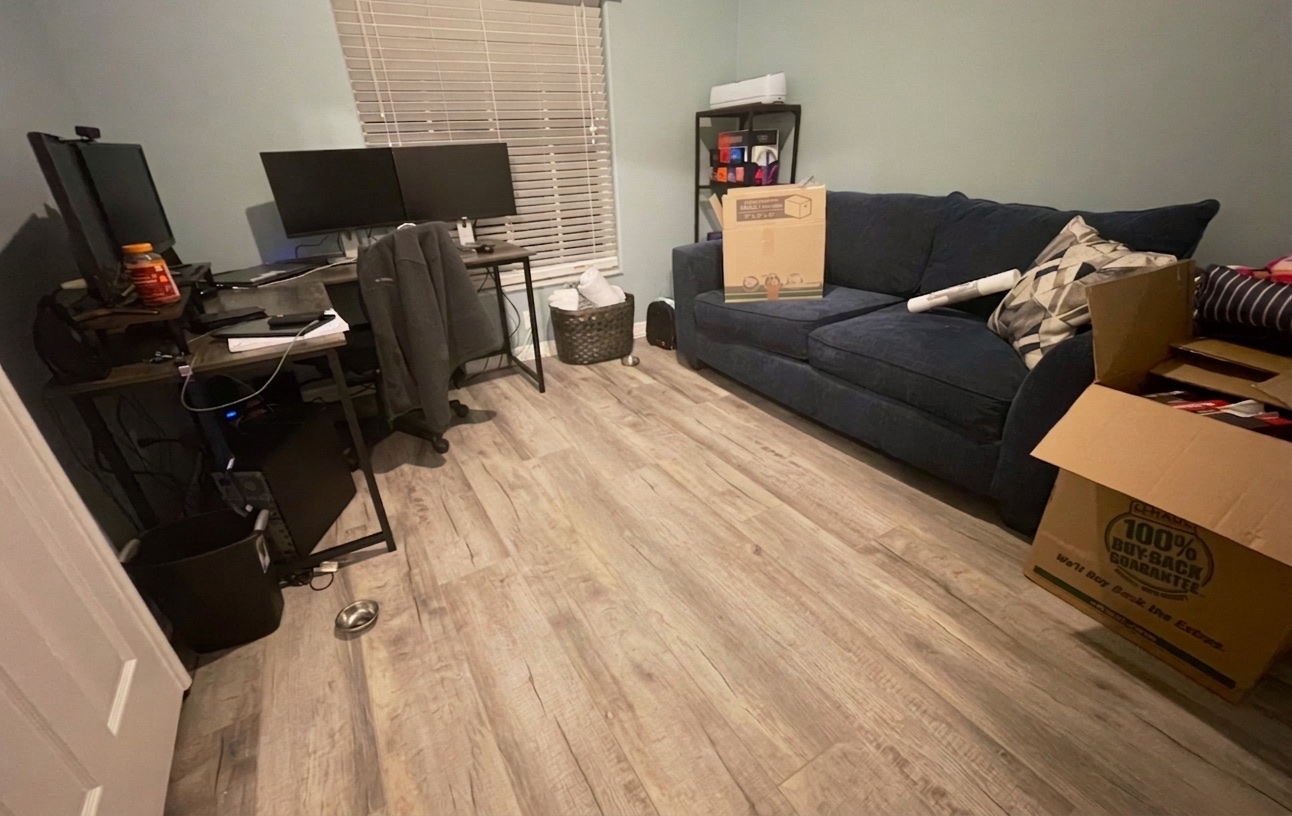} &
      \teaserimg[Room scan 1, SD + LoRA + Controlnet]{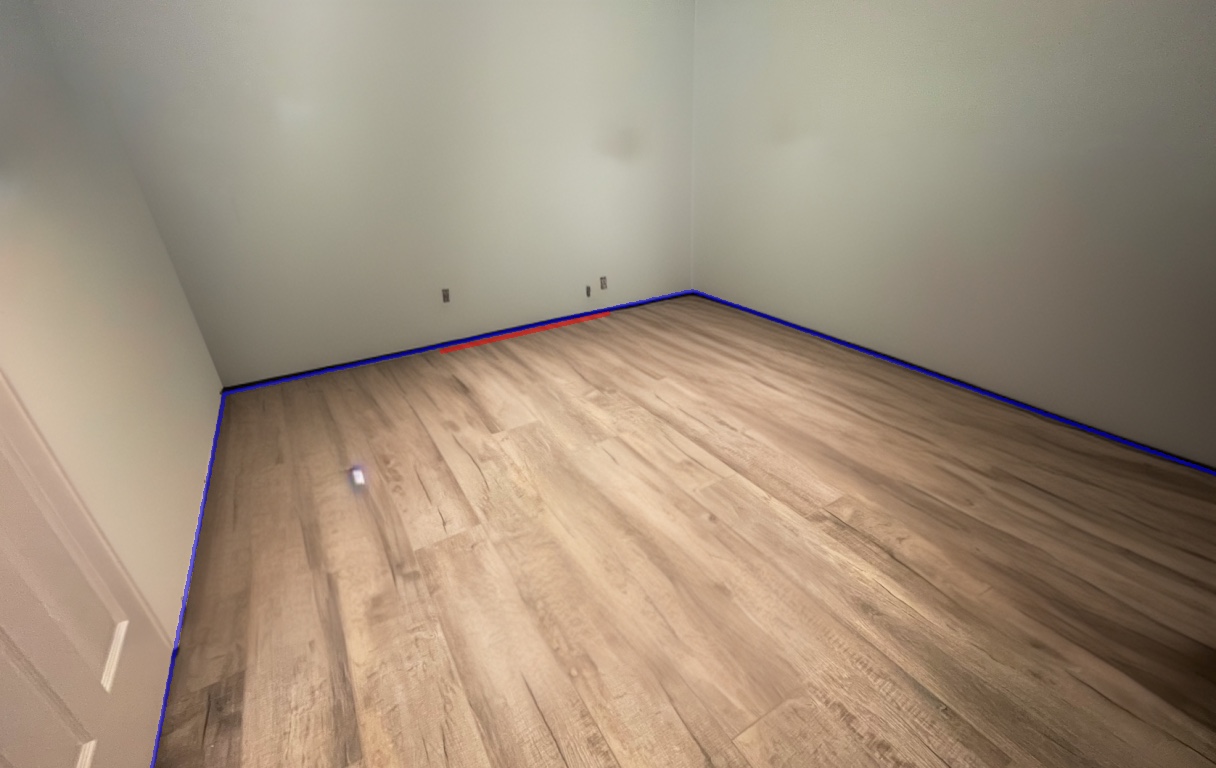} &
      \teaserimg[Room scan 1, FLUX.2]{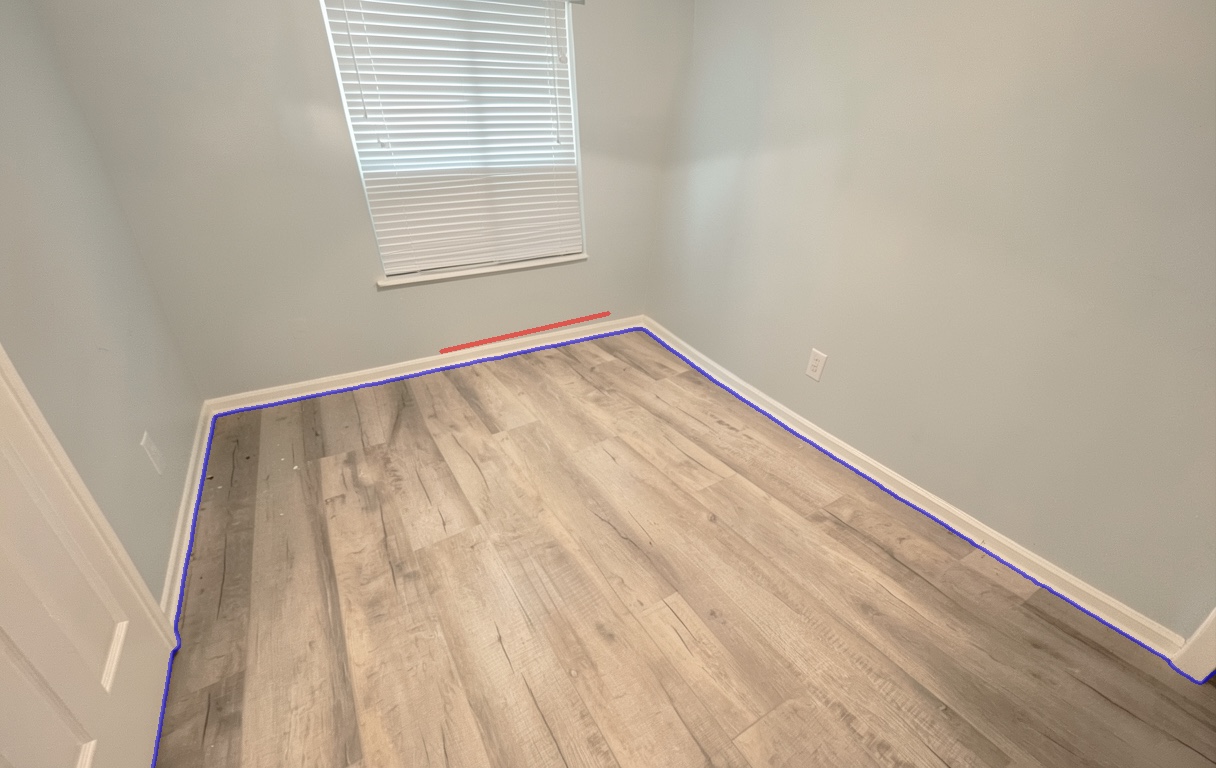} &
      \teaserimg[Room scan 1, FLUX.2 + LoRA (ours)]{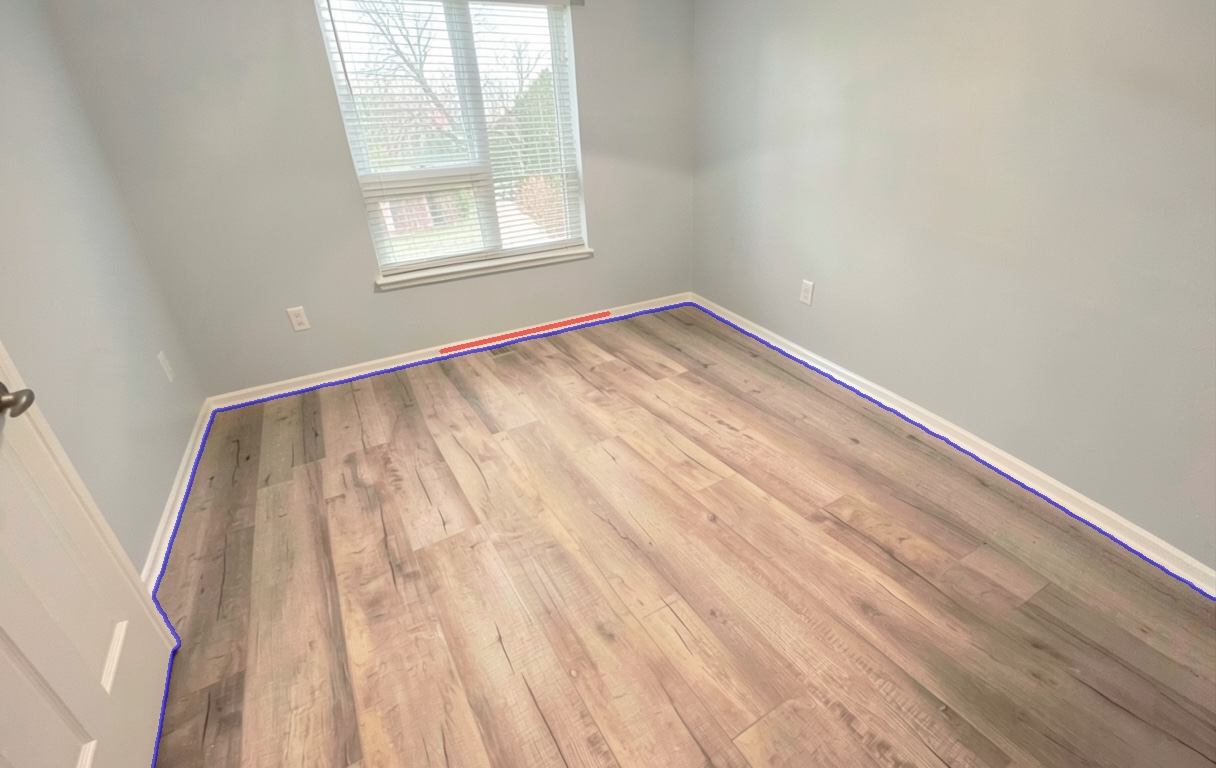} \\[0.5em]
      \teaserimg[Room scan 2, input]{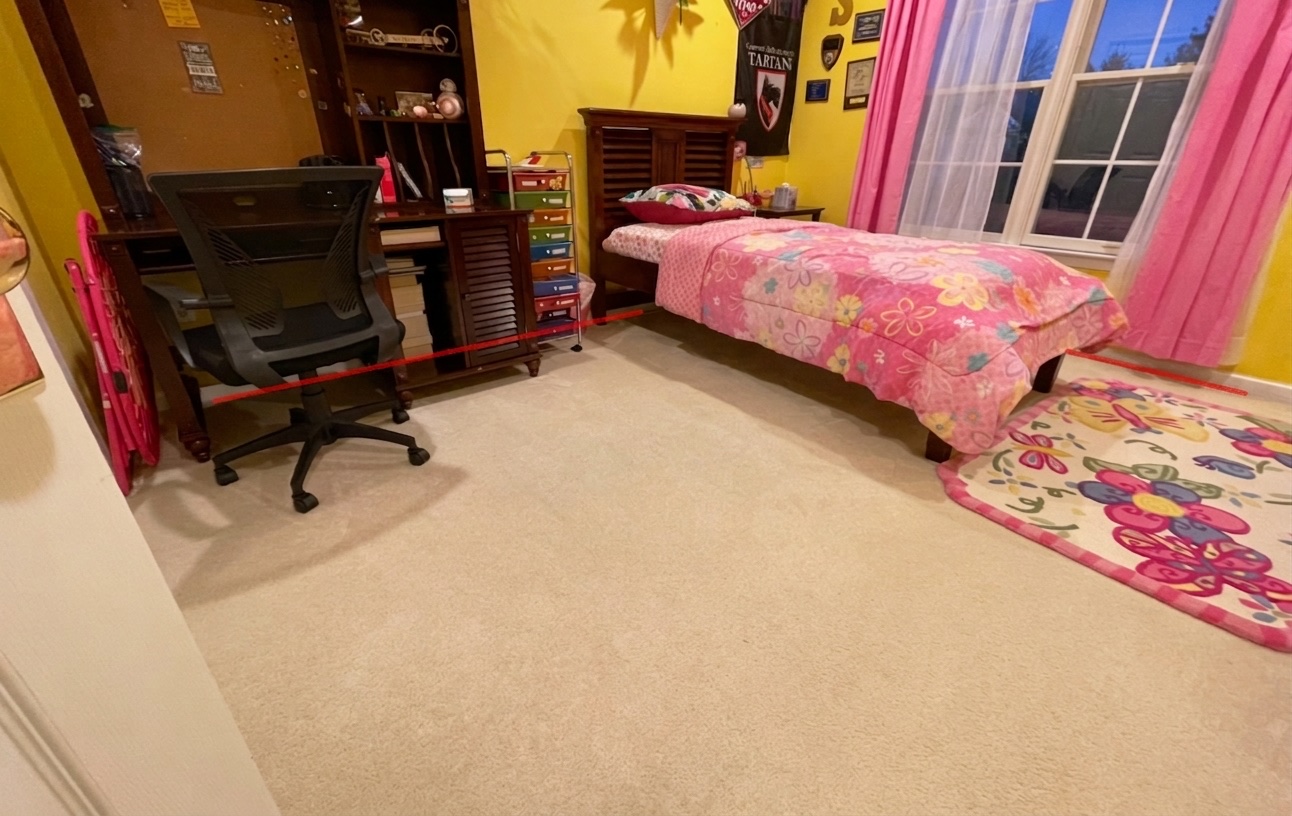} &
      \teaserimg[Room scan 2, SD + LoRA + Controlnet]{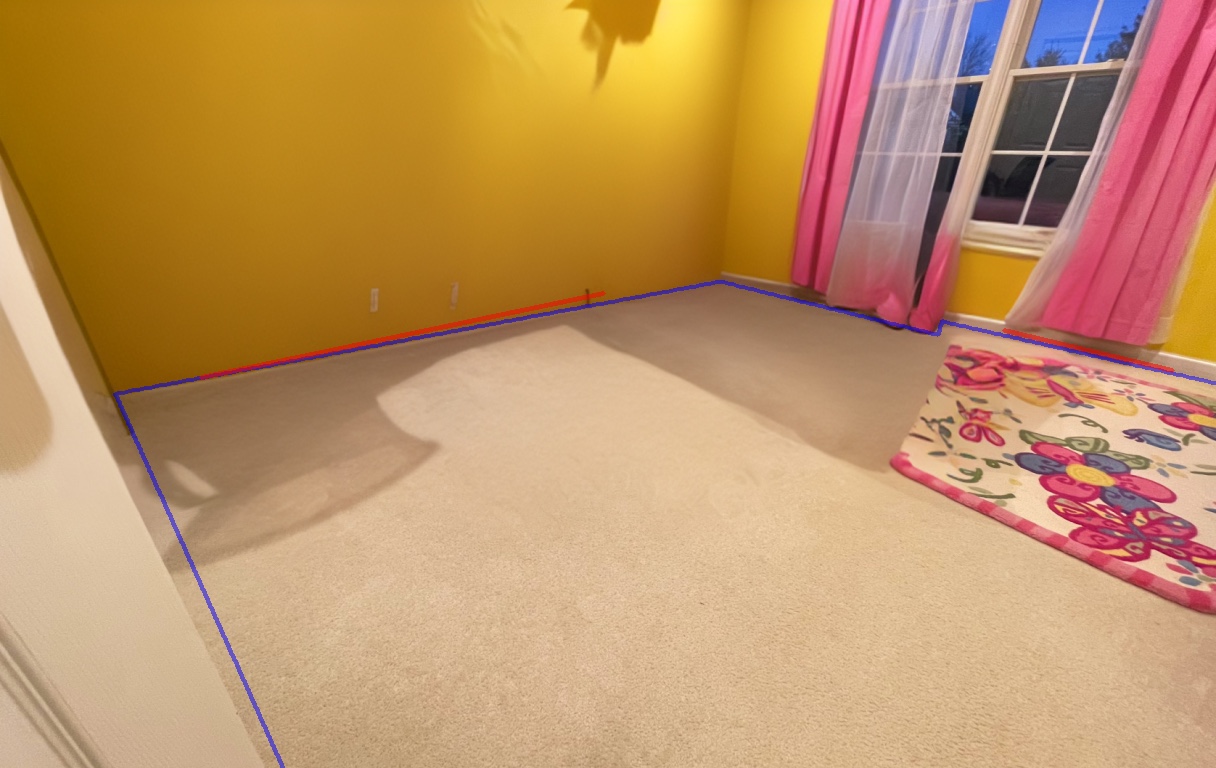} &
      \teaserimg[Room scan 2, FLUX.2]{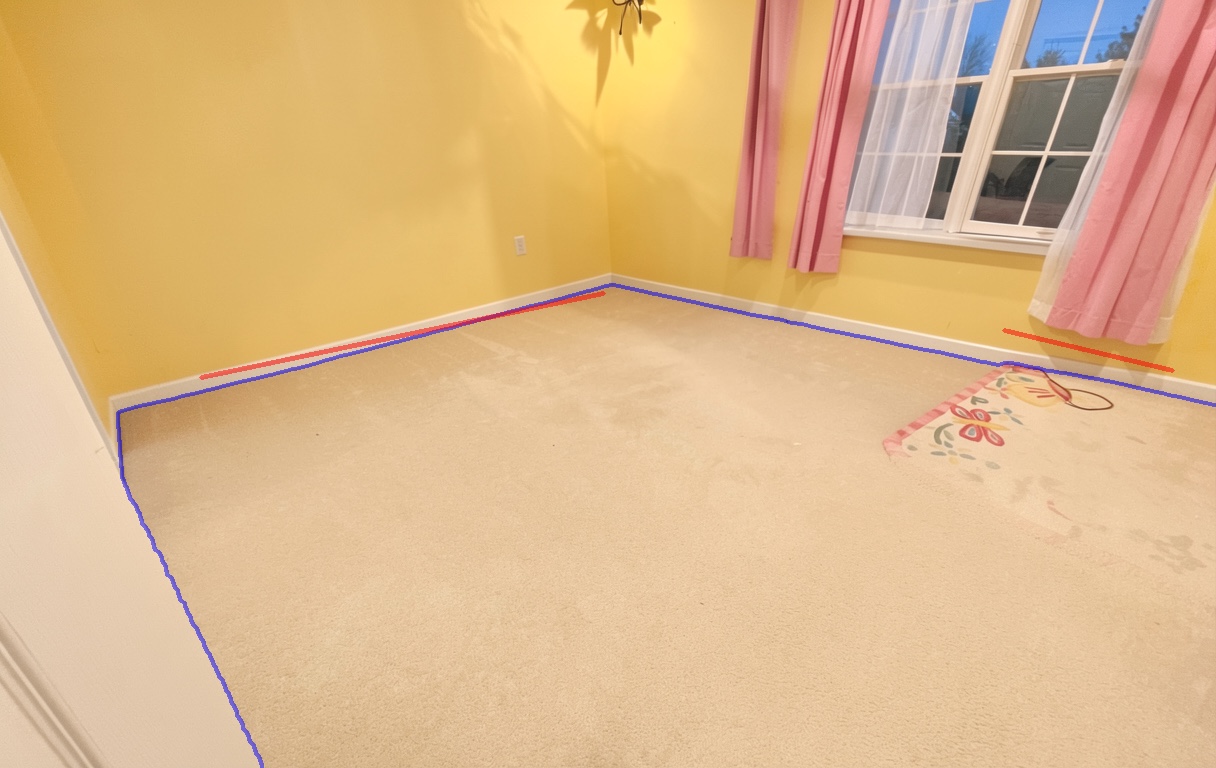} &
      \teaserimg[Room scan 2, FLUX.2 + LoRA (ours)]{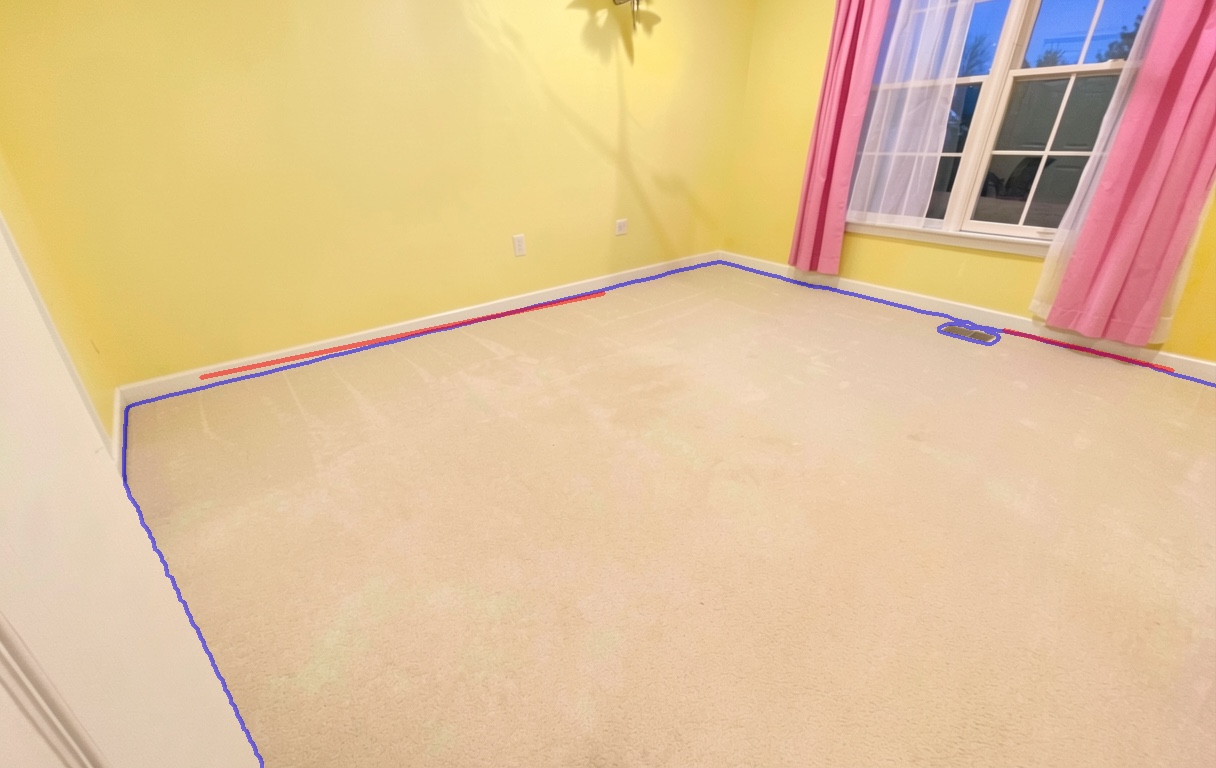} \\
    \end{tabular}%
  };
  \node[anchor=west, draw, fill=white, fill opacity=0.85, text opacity=1,
        rounded corners, inner sep=3pt, font=\footnotesize]
       at ([shift={(3pt,8pt)}]teasergrid.south west) {%
    \begin{tabular}{@{}l@{}}
      \teaserlegendline{red} ground-truth floor\\
      \teaserlegendline{blue} generated floor
    \end{tabular}};
\end{tikzpicture}

  \caption{Qualitative comparison of defurnishing methods on two room scans. Red lines denote ground-truth boundaries; blue lines denote generated floor boundaries. SD+LoRA+ControlNet preserves geometry but sacrifices texture fidelity, whereas FLUX.2 improves realism at the cost of geometry. FLUX.2 + LoRA (ours) achieves both geometric consistency and high-quality synthesis.}
  \label{fig:teaser}
}

\abstract{%
   Erasing all furniture from a single room image to generate its empty counterpart is valuable for XR applications such as diminished reality and virtual staging. The challenge is twofold: inpainting the furniture regions with realistic content, and doing so without distorting the room's 3D geometry. Prompt-driven image editors, such as FLUX.2, handle realism impressively, but their structural faithfulness is inferior to explicit layout conditioned inpainting. To improve geometric consistency, we propose a training recipe for finetuning a LoRA on 2{,}500 furnished-empty room pairs. FLUX.2 with our LoRA adaptor significantly improves geometric consistency while retaining the high texture fidelity. This single-photo inpainter can be an effective building block for depth-aware and 3D-consistent XR pipelines.
}


\begin{document}

\firstsection{Introduction}
\maketitle
\vspace{-0.3\baselineskip}

\begin{figure}[tb]
  \centering
  \newlength{\vstagingcolwidth}
\setlength{\vstagingcolwidth}{0.325\linewidth}
\newcommand{\vstagingimg}[2]{%
  \begin{tikzpicture}
    \node[inner sep=0pt] (img) {\includegraphics[width=\vstagingcolwidth, alt={Room photo, #2}]{#1}};
    \node[anchor=south west, draw, fill=white, fill opacity=0.85, text opacity=1,
          rounded corners, inner sep=2pt, font=\scriptsize]
         at ([shift={(2pt,2pt)}]img.south west) {#2};
  \end{tikzpicture}%
}

\centering
\setlength{\tabcolsep}{2pt}
\begin{tabular}{@{}c@{\hspace{3pt}}c@{\hspace{3pt}}c@{}}
  \vstagingimg{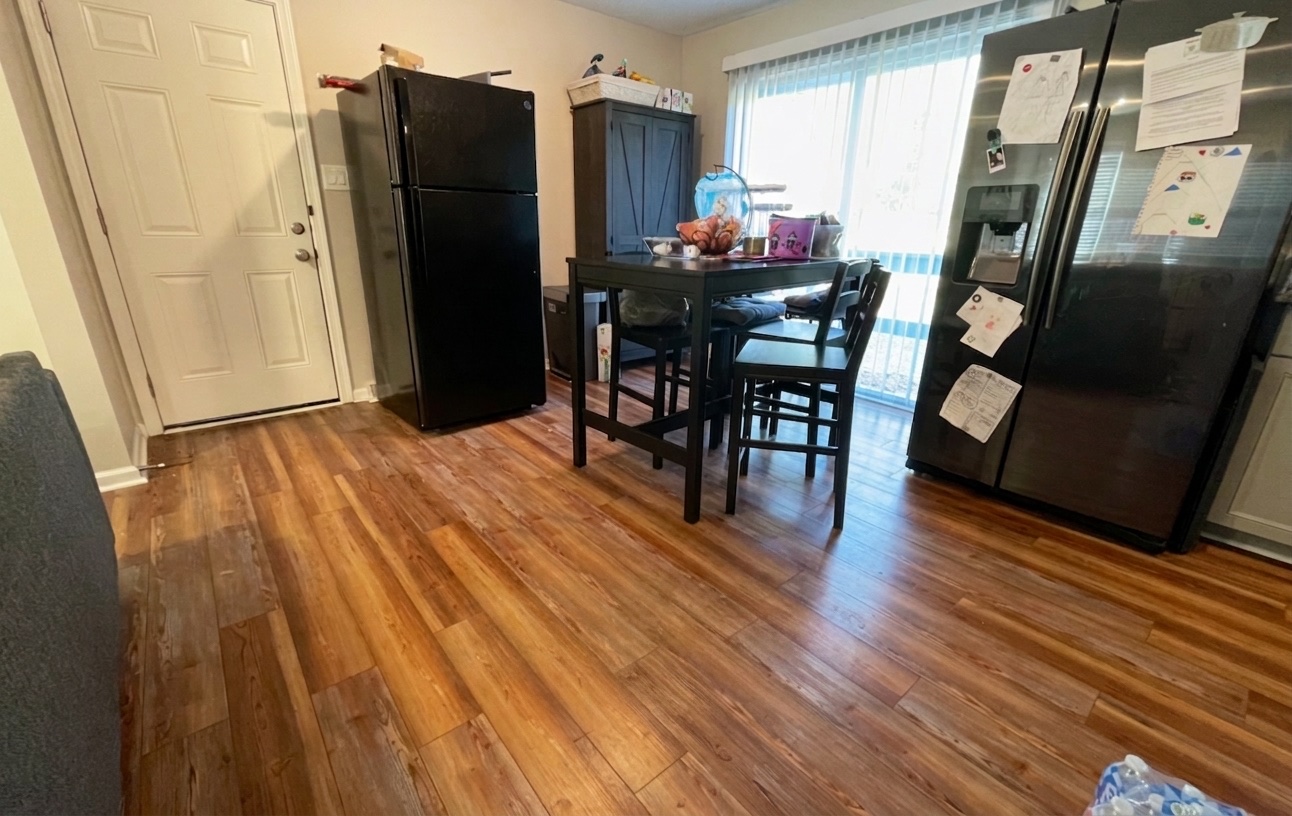}{original} &
  \vstagingimg{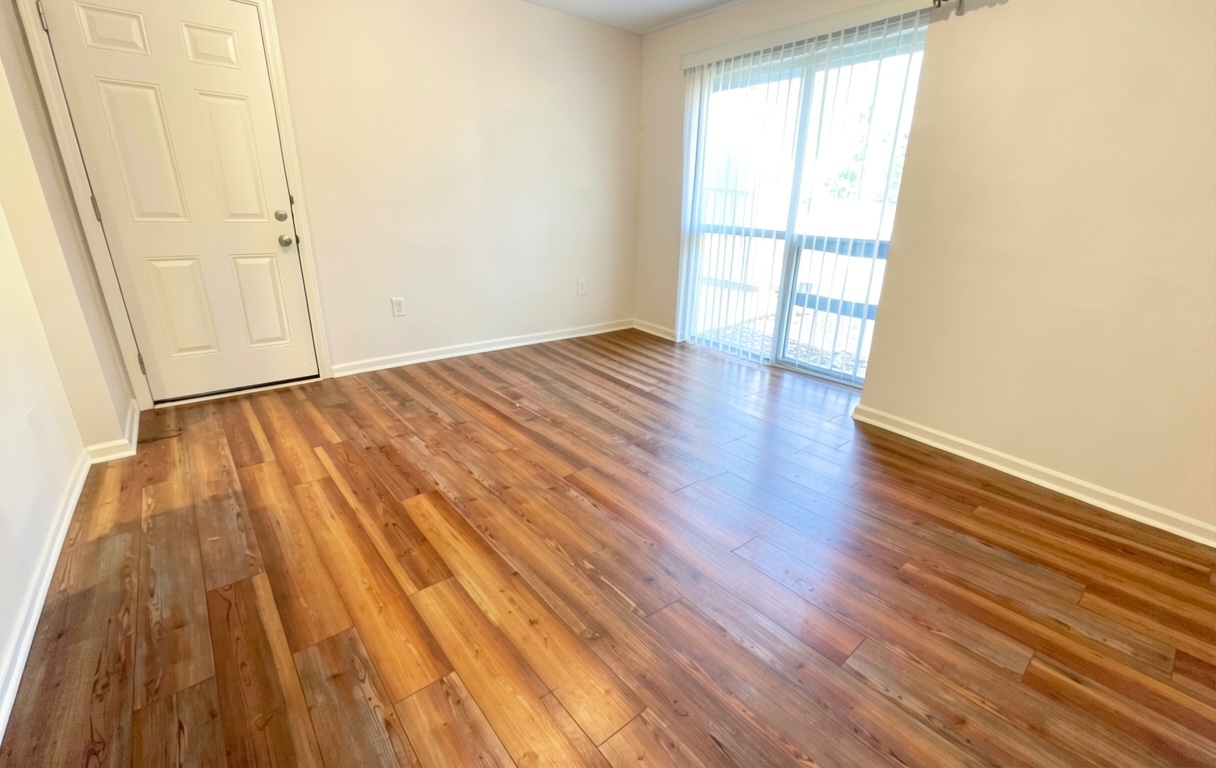}{defurnished} &
  \vstagingimg{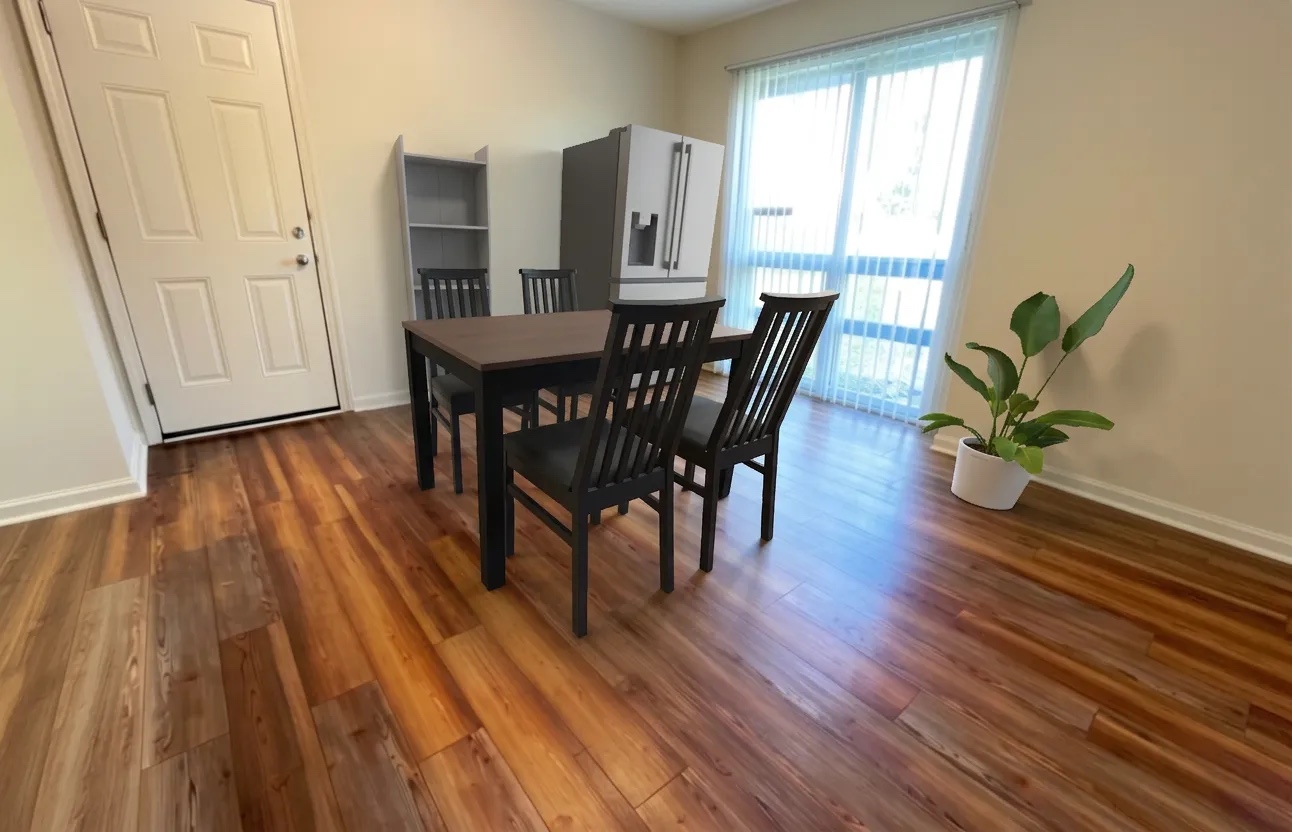}{virtually staged} \\
\end{tabular}

  \caption{Example of defurnishing and virtually staging a room.}
  \label{fig:vstaging}
\end{figure}

Defurnishing is the virtual removal of furniture from a photo, 3D scan, or other digital representation of a room for XR applications such as diminished reality and virtual staging (\cref{fig:vstaging}). Achieving it requires not only inpainting furniture and their shadows, but also preserving geometric consistency so that reconstructed walls and floors remain faithful to the underlying room structure~\cite{kulshreshtha_layout_2022}. Such consistency is particularly important in AR and MR, where virtual objects are anchored to the physical environment. Diminished reality (DR) is an established XR problem that aims to remove or conceal real-world objects while maintaining a coherent scene appearance~\cite{mori2017survey}. Recent DR methods leverage explicit scene geometry through structure-aware RGB-D inpainting~\cite{gsaxner2024deepdr} or by restoring background scene content from pre-captured 3D Gaussian Splat representations~\cite{waldow2025dimsplat}. In contrast, we focus on improving prompt-based editing from a single RGB image. Our lightweight LoRA adaptation can serve as the generative inpainting module within future depth-aware and 3D XR pipelines.

We focus on removing all furniture from a single RGB image to generate an empty-room image. One way to preserve structure is through layout-guided per-plane inpainting~\cite{kulshreshtha_layout_2022}, but it is not robust to shadows and often introduces artifacts in non-planar regions. Prompt-based image editors such as FLUX.2~\cite{bfl_flux2_klein_2026} and Qwen-Image-Edit~\cite{wu2025qwen} produce highly realistic edits but offer no explicit geometric constraints. This raises our central question: \emph{How geometrically consistent are prompt-based editors for defurnishing, and how can we improve them?}

We compare layout-guided Stable Diffusion (SD) inpainting against FLUX.2 using LPIPS~\cite{zhang2018unreasonable} to measure texture fidelity and Layout Error~\cite{koch2018evaluation} to measure geometric consistency. To improve FLUX.2's geometric consistency, we propose a lightweight LoRA finetuning recipe on 2{,}500 furnished-empty room pairs. FLUX.2 with our LoRA adaptor achieves the best texture fidelity and geometric consistency across all three methods, as shown qualitatively in \cref{fig:teaser}.

\section{Method}
\label{sec:method}

\begin{figure}[tb]
  \centering
  \includegraphics[width=0.86\linewidth,
                   alt={Diagram of the three benchmarked inpainting approaches:
                        Stable Diffusion inpainting with LoRA via ControlNets,
                        FLUX.2, and FLUX.2 with LoRA.}]{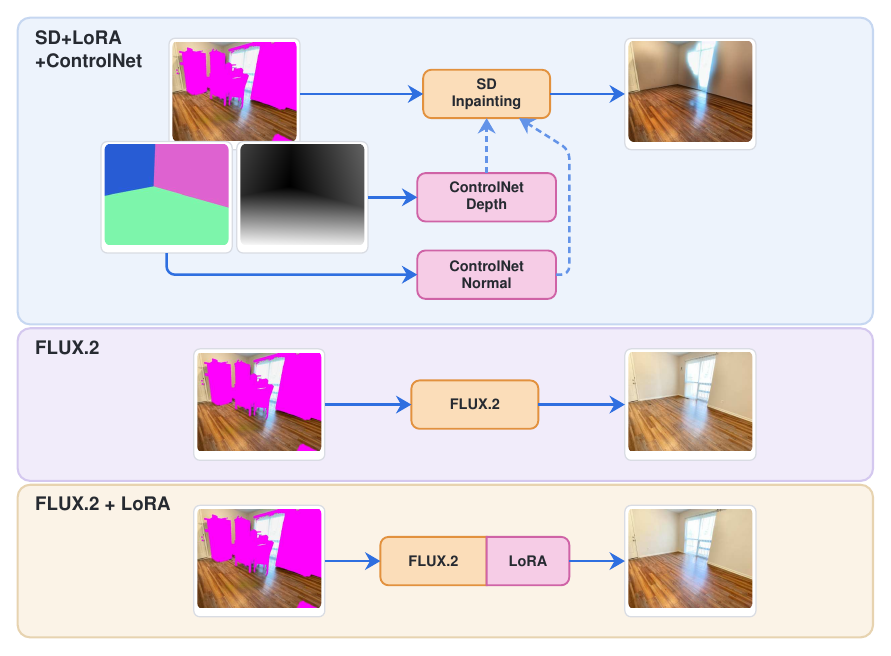}
  \caption{We benchmark three approaches: (1)~Stable
           Diffusion Inpainting + LoRA via ControlNets, (2)FLUX.2, and
           (3)~FLUX.2 + LoRA}
  \label{fig:approaches}
\end{figure}

We compare three inpainting approaches (\cref{fig:approaches}), given as input an RGB image and a binary mask of the furniture regions. The mask can be either user defined or derived using segmentation models. 
For LoRA finetuning, we collect and prepare a dataset of 2{,}500 furnished-empty room pairs in-house.

\subsection{SD + LoRA + ControlNet}
We condition Stable Diffusion 1.5 (SD1.5) inpainting with the room layout. Since the base SD1.5-inpainting often leaves residual furniture shadows, we also finetune a LoRA on our dataset of furnished-empty pairs. Layout is estimated via a 3D layout extractor~\cite{lianos_robust_2022}, which uses perceptual cues such as depth and semantic segmentation to localize the wall and floor planes in 3D.
We render depth and normal maps from this model and feed them to SD1.5-inpainting through two ControlNets~\cite{zhang2023adding}, each at a condition weight of $\alpha=0.5$.

\subsection{FLUX.2}
We apply the mask to the RGB image, marking the masked regions with magenta. We then use the 4-step distilled version of FLUX.2[klein]~\cite{bfl_flux2_klein_2026} on the masked image with the prompt:
\par\vspace{2pt}
{\noindent\small\ttfamily\raggedright
This is a photo of a room in which all objects are marked with a magenta mask. Remove all the masked regions and their shadows from this photo, while ensuring that all the unmarked regions remain absolutely unchanged.\par}
\vspace{2pt}\noindent
With no geometric conditioning, this serves as a reference point for evaluating the structural consistency of a purely prompt-driven editor.

\subsection{FLUX.2 + LoRA}
\label{sec:klein-lora}
To improve geometric consistency, we adapt FLUX.2[klein] by finetuning a LoRA (rank $64$, alpha $32$) on an in-house dataset of 2{,}500 furnished-empty room pairs. Inputs are a masked RGB image and a prompt, same as the previous approach. We train for 3{,}000 steps, with learning rate $1\times10^{-4}$, batch size $4$, height $768$, ema-decay $0.999$. Differential guidance is turned on to encourage focus on inpainted areas. Training is done using ai-toolkit\cite{ostris2025aitoolkit} on a single NVIDIA A5000, and takes 12 hours. During inference, the LoRA is applied onto the distilled 4-step model. 

\subsection{Evaluation metrics}
\label{sec:metrics}
We evaluate the defurnishing pipelines along two criteria: texture fidelity and geometric consistency.
For \emph{texture fidelity} we report the standard perceptual similarity score, LPIPS, on a test-set of 300 furnished-empty room pairs.
For \emph{geometric consistency} we report \emph{Layout Error} on a test-set of 100 furnished room images annotated with layout boundaries. Layout Error is the distance of each annotated layout boundary pixel to the nearest predicted one~\cite{koch2018evaluation}. The predicted layout is the boundary of the `floor' region, extracted from semantic segmentation of the inpainted image.
\begin{equation}
  \mathrm{Layout\ Error} = (1/|G|) \textstyle\sum_{g\in G} \min\big(d(g, P),\, \tau\big),
\end{equation}
where $G$ and $P$ are the ground-truth and predicted edge pixels, $d(g,P)=\min_{p\in P}\|g-p\|$ the distance to the nearest predicted edge, and $\tau$ a threshold capping the penalty for severely missed boundaries.
\section{Results and Conclusion}
\label{sec:results}

\Cref{tab:quantitative} summarizes the performance of the three methods. SD+LoRA+ControlNet achieves lower Layout Error than FLUX.2 due to explicit layout conditioning, but obtains worse LPIPS scores, reflecting the stronger image synthesis capability of FLUX.2. Although FLUX.2 produces realistic edits, it can distort room geometry since it is a generalist model, not optimized for preserving geometry. Our lightweight LoRA adaptation specializes FLUX.2 for defurnishing by learning task-specific geometric cues, including vanishing lines, room element sizes, and scene layout. Consequently, FLUX.2 with our LoRA adaptor achieves the best performance on both texture fidelity and geometric consistency.

\begin{table}[tb]
  \caption{\small Quantitative comparison of inpainting models; all metrics
           defined in ~\cref{sec:metrics} ($\downarrow$: lower is better). Best
           result per column in \textbf{bold}.}
  \label{tab:quantitative}
  \centering\scriptsize
  \setlength{\tabcolsep}{4pt}
  \begin{tabu} to \linewidth {@{}lcc@{}}
    \toprule
    Model & LPIPS $\downarrow$ & \shortstack{Layout Error (px) $\downarrow$}  \\
    \midrule
    SD + LoRA + ControlNet & 0.249 & 1.84 \\
    FLUX.2  & 0.170 & 2.03 \\
    FLUX.2 + LoRA (ours) & \textbf{0.166} & \textbf{1.41} \\
    \bottomrule
  \end{tabu}
\end{table}

To conclude, generalist prompt-based image editors achieve high visual realism but lack geometric faithfulness for room defurnishing, which our lightweight LoRA adaptation recovers while retaining texture fidelity. Our approach is image-only: it uses no explicit 3D conditioning and builds no persistent scene representation, so multi-view and temporal consistency are not guaranteed. Inference takes up to 5 seconds per image, so our method currently supports offline scene preparation rather than real-time XR. Future work will incorporate explicit geometry through depth or normal conditioning, and the resulting model can distill faster real-time variants or precompute empty-room assets for DR systems such as DimSplat~\cite{waldow2025dimsplat}.
\acknowledgments{%
AI assistance was used for figure formatting and proofreading.}

\bibliographystyle{abbrv}
\bibliography{geomagical-labs}
\end{document}